\documentclass[journal]{IEEEtran}
\usepackage[pdftex]{graphicx}
\usepackage{times}
\usepackage{epsfig}
\usepackage{amsmath}
\usepackage{amssymb}
\usepackage{multirow}
\usepackage{cite}
\usepackage{subfigure}
\usepackage{algorithmic}
\usepackage{ifpdf}
\usepackage{array}
\usepackage{color}
\usepackage[ruled,linesnumbered]{algorithm2e}
\usepackage{algorithmic}

\begin{document}
%
\title{Learn to Adapt for Monocular Depth Estimation}
%
%
%

\author{Qiyu~Sun, Gary G. Yen,~\IEEEmembership{Fellow, IEEE},
        Yang~Tang,~\IEEEmembership{Senior~Member,~IEEE,}~and Chaoqiang~Zhao
\thanks{ Q. Sun, Y. Tang, and C. Zhao are with the Key Laboratory of Smart Manufacturing in Energy Chemical Process, Ministry of Education, East China University of Science and Technology, Shanghai, 200237, China (e-mail: yangtang@ecust.edu.cn (Y. Tang)).}
\thanks{ Gary G. Yen is with the School of Electrical and Computer Engineering,
	Oklahoma State University, Stillwater, OK 74075 USA (e-mail: gyen@okstate.edu).}
}

\maketitle

\begin{abstract}
Monocular depth estimation is one of the fundamental tasks in environmental perception and has achieved tremendous progress in virtue of deep learning. 
However, the performance of trained models tends to degrade or deteriorate when employed on other new datasets due to the gap between different datasets.
Though some methods utilize domain adaptation technologies to jointly train different domains and narrow the gap between them, the trained models cannot generalize to new domains that are not involved in training.
To boost the transferability of depth estimation models, we propose an adversarial depth estimation task and train the model in the pipeline of meta-learning.
Our proposed adversarial task mitigates the issue of meta-overfitting, since the network is trained in an adversarial manner and aims to extract domain invariant representations.
In addition, we propose a constraint to impose upon cross-task depth consistency to compel the depth estimation to be identical in different adversarial tasks, which improves the performance of our method and smoothens the training process. 
Experiments demonstrate that our method adapts well to new datasets after few training steps during  {the test procedure}.
\end{abstract}


%
\IEEEpeerreviewmaketitle

\section{Introduction}

Monocular depth estimation is a classical task in computer vision, which aims to estimate the distances between the objects in environment and the agent itself~\cite{Zhou2017Unsupervised,sun2021unsupervised}, and thus it is an essential task in environmental perception~\cite{yuan2015scene,tang2020overview}. Recently, 
deep learning-based depth estimation methods, including supervised methods~\cite{Eigen2014Depth}  and unsupervised methods~\cite{Zhou2017Unsupervised, sun2021unsupervised,zhao2020masked}, are proposed and  achieve significant progress.
The supervised methods are trained through images with ground truth and their performances are often reliable. On the other hand, the unsupervised methods are trained through unlabeled images and hence the training data are more available. 
In general, though the deep learning-based depth estimation models perform well on the training dataset, they suffer from distinct performance degradation when applied to new datasets~\cite{zhang2020autonomous}. Such degradation occurs due to the differences of data distributions exist between the training and testing datasets, and such phenomenon is known as domain shift~\cite{csurka2017domain, shao2014transfer}.
In order to ameliorate the issue of domain shift, some domain adaptation technologies are proposed to fill the gap existing between different datasets~\cite{csurka2017domain,wilson2020survey}. For example, some approaches attempt to reduce the domain gap between synthetic datasets and real datasets in depth estimation~\cite{bozorgtabar2019syndemo,vankadari2020unsupervised,nath2018adadepth,kundu2019adapt}. 
Even though, synthetic (source domain) and real datasets (target domain) are indispensable during training, and the model cannot generalize to new domains that not participating in training.
In practical scenarios, the model that can adapt to various unseen domains is urgently needed.


%
%
%
%
%

\begin{figure}[t]
	\centering
	\subfigure[\scriptsize{RGB image}]{
		\includegraphics[width=0.2\textwidth]{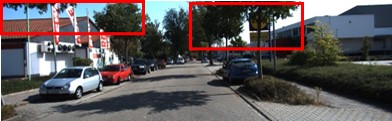}
	}
	\quad
	\subfigure[\scriptsize{Ground truth}]{
		\includegraphics[width=0.2\textwidth]{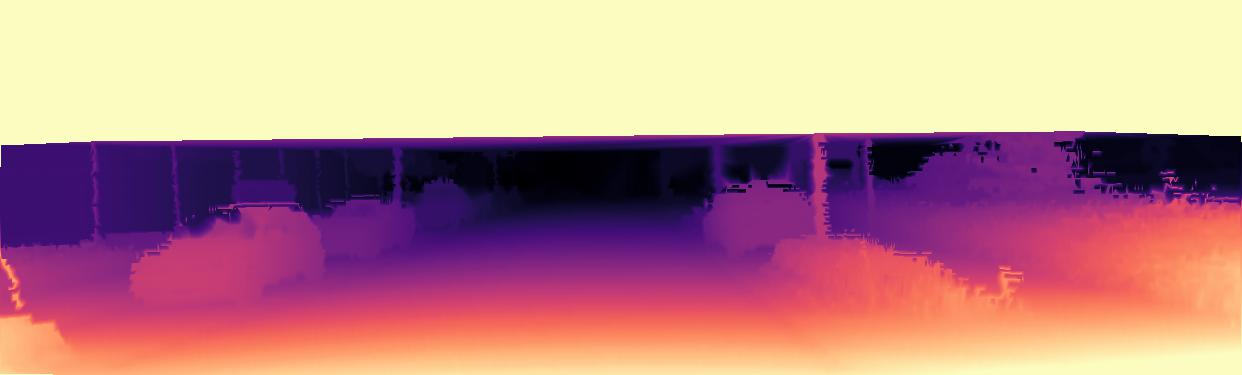}
	}
	\quad
	\subfigure[\scriptsize{Monodepth2~\cite{2019digging} (20 epochs)}]{
		\includegraphics[width=0.2\textwidth]{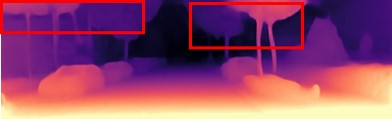}
	}
	\quad
	\subfigure[\scriptsize{Monodepth2~\cite{2019digging} (0.5 epoch)}]{
		\includegraphics[width=0.2\textwidth]{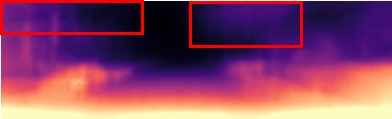}
	}
	\quad
	\subfigure[\scriptsize{Monodepth2~\cite{2019digging} + MAML~\cite{finn2017model} (0.5 epoch)}]{
		\includegraphics[width=0.2\textwidth]{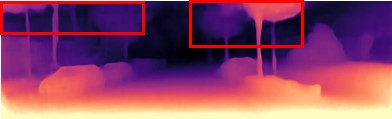}
	}
	\quad
	\subfigure[\scriptsize{Our method (0.5 epoch)}]{
		\includegraphics[width=0.2\textwidth]{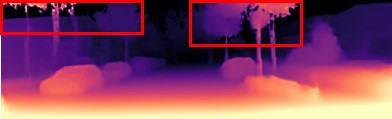}
	}
	\caption{The comparison of depth estimation results. 
				(a) The input image; (b) Depth ground truth; 
		(c)-(d) Results obtained by Monodepth2~\cite{2019digging} trained for 20 epochs and 0.5 epoch, respectively; (e) Result obtained by Monodepth2 trained in MAML pipeline directly for 0.5 epoch; (f) Result of our proposed  trained for about 0.5 epoch. Our method is better at reconstructing the detail of objects and providing sharper predictions, for example, the outline of trees, than Monodepth2 within just few steps during  {the test procedure}.	
}
	\label{introduction}
\end{figure}


Recently, meta-learning~\cite{schmidhuber1987evolutionary,andrychowicz2016learning}, also known as learning to learn, shows the possibility of learning an adaptive model which can generalize well to new domains with limited training steps. 
 {In the training phase,} the model is trained on different tasks to learn how to adapt to new tasks or new domains with few samples.  {In the test phase, the model is trained with new datasets on the basis of the model obtained in the training phase.}
Since the model trained with meta-learning includes prior mutual knowledge about training tasks, it can be applied to a new task with a small amount of follow-up training steps  {during test}.
It is worth noting that large quantities of datasets are an essential precondition for the excellent transferability of meta-learning, therefore meta-learning is well applied to image classification tasks thanks to the existing numerous classification datasets~\cite{lai2020learning,1}. 
If the tasks for  {training} are limited, meta-learning suffers from distribution shift issue as well, which is also called the \textit{meta-overfitting}~\cite{tian2021consistent,1,2}.
This is because that the limited tasks cannot ensure the distribution of testing tasks resembles that of training tasks.
Unfortunately, the datasets for depth estimation are quite limited, which raises challenges when integrating meta-learning into depth estimation.

As the problem of poor generalization between datasets imposes great  restrictions on real-world applications of depth estimation, this work explores the possibility of learning a more adaptive depth estimation model which can generalize well to new domains.
We refer to model-agnostic meta-learning (MAML)~\cite{finn2017model}, a well-known 
algorithm of meta learning, to endow depth estimation models the transferability across different domains. 
To maintain the generalization of meta-learning in depth estimation tasks in case of limited training datasets, we design an adversarial depth estimation task. 
Our proposed adversarial task estimates the depth maps of two images sampled from different datasets simultaneously and such combination of datasets increases the amount of tasks. Then, the estimated depth maps are compared in an adversarial manner to extract the domain invariant representations, thus avoiding the overfitting in  {training}.
Since the supervised methods rely on plenty of training data with depth ground truth, which is laborious and time consuming to collect for real-world scenes, we adopt the unsupervised pipeline using monocular image sequences for training  to obtain more training datasets~\cite{Zhou2017Unsupervised,2019digging}. 
Both real images collected by cameras and synthetic images generated by graphics engines~\cite{gaidon2016virtual,ros2016synthia} are utilized for training  in this work.
In view of the fact that the unsupervised pipeline is vulnerable to the dynamics in environment~\cite{sun2021unsupervised}, we propose a constraint to compel the depth maps estimated in different adversarial tasks to be identical. We name it as the cross-task depth consistency constraint and it is applied in meta-update. In addition, the proposed constraint helps to improve the performance of  {training} and smoothen the convergence curve because it establishes the connection between different  {training} tasks and then the model is optimized in a collaborative manner.

To sum up, we propose an adversarial domain-adaptive algorithm for monocular depth estimation, which is able to generalize to unseen datasets or scenes quickly in test phase and achieves a satisfactory performance. The adversarial depth estimation task we introduce alleviates the meta-overfitting even trained on a handful of datasets. Experiments demonstrate that our meta-learning based monocular depth learning method can rapidly adapt to new, unseen datasets during  {test}. 
As shown in Figure \ref{introduction}, our method (Figure 1(f)) achieves better performance than the state-of-the-art work~\cite{2019digging} (Figure 1(c))  after few steps of training. 
For further comparison, we train the model using the basic ~\cite{2019digging} (Figure 1(d)) and in the pipeline of  MAML (Figure 1(e)) for 0.5 epoch, respectively. The results show that our method adapts well to a new dataset rapidly.
Our contributions are summarized as follows:



\begin{itemize}
	\item [-] We propose an adversarial training pipeline for monocular depth estimation based on meta-learning, which learns appropriate initial network  parameters for adapting efficiently to unseen domains with few steps of adaptation. Meanwhile, our method alleviates the issue of meta-overfitting even when trained on few datasets.

	\item [-] We propose a cross-task depth consistency constraint handling for meta-update. It compels the depth estimated from the same image in different tasks of meta-learning to be identical, which stabilizes the training process and improves the performance of our method.	
	
	\item [-] The model trained by our method adapts well to  datasets which do not appear in the  {training} phase. Our method trained through few updates obtains comparable results with some state-of-the-art works trained with much more time.

\end{itemize}

The organization of this paper is arranged as follows. Section II presents a brief review of related works. The method of this paper is introduced in the Section III. The Section IV shows the experimental results about the transferability of the proposed depth estimation model. Finally, the Section V draws a conclusion on this study and some future prospects are provided.

\section{Related Works}

\subsection{Depth Estimation}
Depth estimation is of great importance in environmental perception, and  many exsiting methods attempt to solve it in an end-to-end manner thanks to deep learning~\cite{zhao2020monocular}.
The methods estimate depth in either a supervised manner or an unsupervised fashion according to training data. 
The supervised methods~\cite{Eigen2014Depth} are trained with images and their corresponding ground truth depth, thus the estimation results are more reliable. 
However, the acquisition of depth labels in real environments is time consuming and costly.
For unsupervised methods, they can be further divided into two groups according to training data, trained on stereo image pairs~\cite{Garg2016Unsupervised,Godard2017Unsupervised} or  monocular image  sequences~\cite{Zhou2017Unsupervised,2019digging,johnston2020self,sun2019cycle}. 
Among all the training modes, the unsupervised methods based on monocular images are the most attractive because plenty of training data are available. 
However, unsupervised methods are not as reliable as supervised ones and are
more vulnerable to the dynamic variations in environments~\cite{sun2021unsupervised}. 
Though all these deep learning based depth estimation methods dedicate to improve the performance of the model on a particular dataset, they neglect the transferability of the model, which imposes significant  restrictions on real-world applications.

\subsection{Adversarial Domain Adaptation}
Recently, deep learning flourishes in the performance of numerous computer vision tasks~\cite{fernando2013unsupervised,hoffman2016fcns,zheng2018t2net}. However, the model trained on a specific dataset is frequently incapable of generalizing well on another dataset due to dataset bias~\cite{torralba2011unbiased, tommasi2017deeper}. 
Domain adaptation ~\cite{csurka2017domain,wilson2020survey} is proposed to narrow the gap between different datasets and proved to be efficient in different computer vision tasks, including classification~\cite{fernando2013unsupervised},  semantic segmentation~\cite{hoffman2016fcns}, and depth estimation~\cite{zheng2018t2net}, etc.


Among various  domain adaptation works, adversarial-based domain adaptation methods are the dominant ones~\cite{csurka2017domain,wilson2020survey}.
Some methods narrow the distribution shift between different datasets in  latent feature space~\cite{tzeng2015simultaneous,ganin2015unsupervised,hoffman2016fcns}, which are categorized as feature-based domain adaptation methods.
Though these feature-based methods can perform well in classification tasks, they tend to fail in more complicated tasks like  depth estimation and semantic segmentation~\cite{liu2018collaborative,zhang2021multitask} . 
The development of Generative Adversarial Networks (GAN)~\cite{goodfellow2014generative} promotes the emergency of  input-based~\cite{kim2020learning,zheng2018t2net,zhao2019geometry}  and output-based domain adaptation methods~\cite{tsai2019domain,wang2020classes,nath2018adadepth,kundu2019adapt}. 
These methods greatly enhance the transferability of deep models among different datasets. The input-based domain adaptation methods almost take advantage of style transfer networks~\cite{zhu2017unpaired,isola2017image} to make the training images indiscernible across domains.  The output-based domain adaptation methods~\cite{tsai2019domain,wang2020classes,nath2018adadepth,kundu2019adapt} take advantage of the network predictions. For pixel wise estimation tasks such as depth estimation, the output predictions contain structured  spatial information, which are not sensitive to domains.

Specially, numerous domain adaptation works focus on depth estimation.
Some methods use an adversarial objective to distinguish which domain the extracted feature comes from with
 feature-based methods~\cite{bozorgtabar2019syndemo,vankadari2020unsupervised}, while other works dedicate to improving the transferability of the model with input-based methods~\cite{atapour2018real,zheng2018t2net,zhao2019geometry,cheng2020s}. Notice that depth estimation is a spatially-structured prediction task, Kundu \textit{et al.}~\cite{nath2018adadepth,kundu2019adapt} improve its across-domain generalization capability by adding a consistency constraint on the output space.
Though domain adaptation methods can relax the domain shift issue between two or several datasets, the model trained using these methods can only perform well on these datasets and tends to fail in other datasets which are not involved in training.

\subsection{Meta-Learning}
Meta-learning~\cite{schmidhuber1987evolutionary,andrychowicz2016learning} is designed to train a model through various tasks to equip the model with the ability of solving new tasks with a small quantity of training samples.
It has gained tremendous attention in few-shot learning~\cite{ravi2016optimization} and reinforcement learning~\cite{wang2016learning}. Recently, meta-learning is applied to some computer vision tasks as well~\cite{zhang2020online,choi2020scene,soh2020meta}.
MAML~\cite{finn2017model} is a well-known 
algorithm for meta learning. 
It prompts the network
to learn  common initial parameters  across different training tasks, and the initial parameters can be reloaded in the testing phase to ensure an advanced generalization performance on unseen tasks quickly. 

However, the superior performance of meta-learning in generalization is obtained by training on various tasks and an appropriate task distribution~\cite{1,2}. 
Some reinforcement learning works have noticed such problem and attempted to address it~\cite{1,2}. 
Mehta \textit{et al.}~\cite{2} utilize domain randomization methods to optimize the distribution of tasks and obtain more stable initial parameters for variable testing tasks.
Lin \textit{et al.}~\cite{1}  propose AdMRL to optimize the worst performance over all the tasks and improve the generalization ability of the model.
The same problem exists in depth estimation as well. Dozens of tasks are needed for training in meta-learning generally~\cite{hospedales2020meta}, while the datasets for depth estimation are quite limited. 
Thus, we train the depth estimation model in an adversarial manner using only four datasets  to extract domains invariant representations, and attempt to alleviate the issue of meta-overfitting in depth estimation tasks due to the limited training tasks.
%


\section{Proposed Method}

In this section, we introduce the proposed method in detail. The adversarial depth estimation task for meta-learning is introduced first in Subsection III-A,  the parameter update of the proposed task is introduced in Subsections III-B, and the proposed cross-task depth consistency constraint is presented in Subsections III-C.

\subsection{Adversarial Task for Depth Estimation}
\label{sec:adversarial}

\begin{figure}[t]
	\centering
	\includegraphics[scale=0.45]{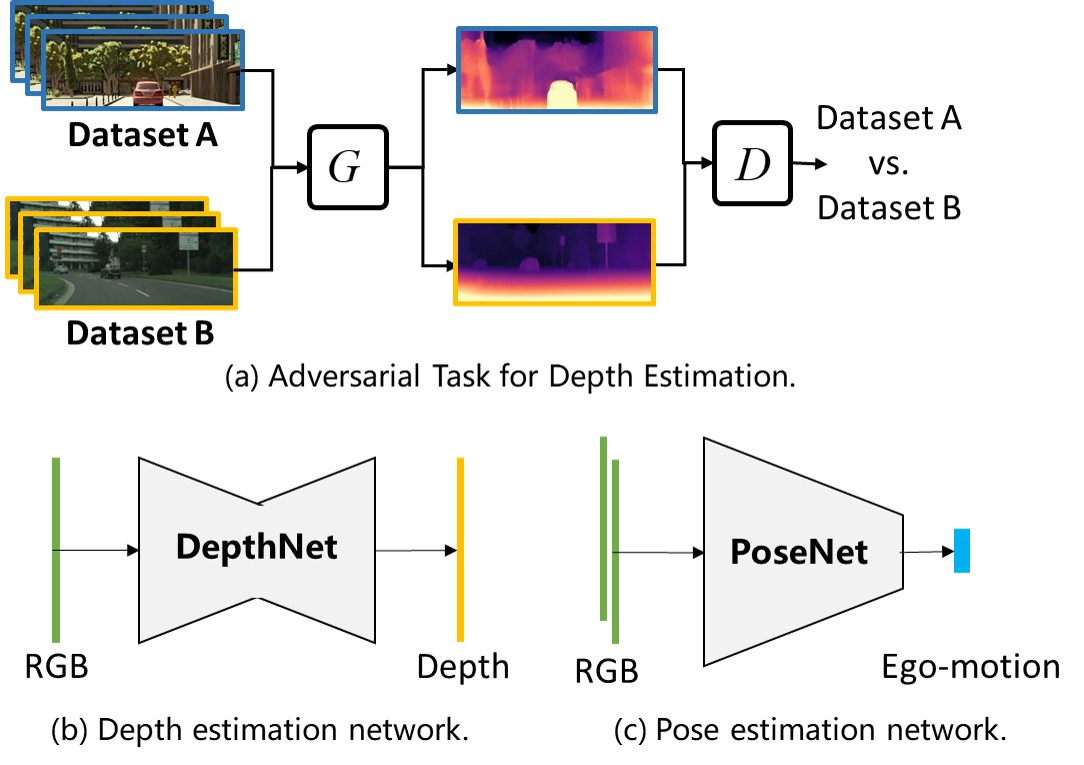}
	\caption{The illustration of the adversarial task designed for meta-learning. (a) The task includes a generator $G$ for depth estimation and a discriminator  $D$ for domain invariant feature extraction. Images from different domains are trained jointly in an adversarial manner. Monocular depth is estimated through sequence of snippets consisting of two consecutive frames through (b) a DepthNet for pixel-level depth estimation and (c) a PoseNet for relative pose estimation between two images.
	}
	\label{UDA}
\end{figure}

As shown in Figure \ref {UDA}(a), the adversarial task designed includes a generator  for monocular depth prediction and a discriminator for domain invariant feature extraction. 
To obtain more training data, we adopt the unsupervised monocular depth estimation pipeline~\cite{2019digging} as the generator.  It includes a DepthNet (Figure \ref {UDA}(b)) for  pixel-level depth prediction and a PoseNet (Figure \ref {UDA}(c)) for  relative pose estimation between two images. During training, the DepthNet takes a single image as input and the PoseNet takes two consecutive images split from a monocular sequence as input. The supervised signal comes from view synthesis~\cite{Zhou2017Unsupervised}. For two consecutive images $I_{a}$ and $I_{b}$ in a monocular sequence, the view synthesis implies that we can synthesize image $I_{a}$ when given the estimated depth map of $I_{a}$, the predicted ego-motion of camera from scene $I_{a}$ to $I_{b}$, and the camera internal matrix.
In addition, since both of our adversarial depth estimation task and MAML used in our method are model-agnostic, our generator can also adopt other existing depth estimation methods.

We train the depth estimation model along with a discriminator to narrow the domain gap existing between different datasets. 
Since depth estimation is a pixel-level regression task and the predicted depth maps contain a wealth of spatially structured information~\cite{nath2018adadepth,kundu2019adapt}, we add the discriminator on the output space to extract domain invariant representations.
Even though adding an additional style transfer network to make training data indiscernible across domains is also effective~\cite{zheng2018t2net,zhao2019geometry,cheng2020s}, we do not employ it in this work to reduce computation complexity. 
That is to say, the depth maps are predicted in pairs and sent to the discriminator, and the discriminator tries to  distinguish which domain  the predicted depth comes from.  
With our designed adversarial depth estimation task, the model can extract domain invariant representations~\cite{tzeng2015simultaneous} to ease the meta-overfitting when trained on few tasks. 

In order to demonstrate the effectiveness of our meta-learning algorithm when trained on only few datasets, we merely choose four representative datasets for  {training} in this work. 
Other datasets for depth estimation can be added for training as well, which may facilitate meta-learning but increase training time. 
To be specific, we choose SYNTHIA~\cite{ros2016synthia}, Cityscapes~\cite{cordts2016cityscapes}, Oxford RobotCar~\cite{maddern20171} and vKITTI~\cite{gaidon2016virtual} as training datasets. 
Two of them are synthetic datasets and the other two are real datasets collected by camera. 
We combine them in pairs (SYNTHIA and Cityscapes, for example) to obtain six different combinations and constitute six tasks used for  {training}, thus this kind of combination increases the number of training tasks.
Therefore, the adversarial task for depth estimation can not only extract domain invariant representations but also increase the number of training tasks, which help to overcome the difficulty in meta-learning brought by the lack of training datasets.



The depth estimation model is trained on both real monocular image sequences and  synthetic image-depth pairs.
Real images without the corresponding ground truth are trained in unsupervised manner, and the mainly supervised signal comes from view reconstruction~\cite{2019digging}. The loss function is denoted as  $\mathcal{L}^{u}$ and it is the same as the loss function used for monocular depth estimation in~\cite{2019digging}.
For synthetic images with ground truth, they are  trained in both of supervised and unsupervised manners. The supervised loss $\mathcal{L}^{s}$ is calculated by the L1-norm of the predicted depth and depth ground truth, and the loss function is $\mathcal{L}^{s} + \mathcal{L}^{u}$.
Then, the loss for depth estimation (generator), $\mathcal{L}^{G}$, is 
\begin{equation}
\mathcal{L}^{G}=\left\{
\begin{aligned}
&\mathcal{L}^{u},  &\text{for real data}  \\
&\mathcal{L}^{s} + \mathcal{L}^{u},   &\text{for synthetic data}  
\end{aligned}
\right. \;.
\end{equation}

%
%
%
%
%

After the training snippets from different datasets being fed into the generator for depth estimation, the depth estimation results are sent into the discriminator. If the discriminator cannot distinguish whether the depth estimation results come from dataset $A$ or dataset $B$, it demonstrates that the generator can extract the domain invariant features. We adopt the GAN loss~\cite{goodfellow2014generative} as the adversarial objective, $\mathcal{L}^{D}$,  
\begin{equation}
\label{eq:gan_loss}
\begin{aligned}
\mathcal{L}^{D}=&\mathbb{E}_{x_{A} \sim p_{\rm data}(x_{A})}[{\rm log}D(G(x_{A}))] \\
&+\mathbb{E}_{x_{B} \sim p_{\rm data}(x_{B})}[{\rm log}D(1-D(G(x_{B})))]\;,
\end{aligned}
\end{equation}
where $x_{A}$, $x_{B}$ are the training data from dataset A and B, $G(x_{A})$ and $G(x_{B})$ aim to estimate the depth from images from dataset A and B, and $D$ tries to distinguish whether  the input depth map is from dataset A  or B.

\subsection{Learn to Adapt with Meta-learning}

Numerous tasks are needed for meta-learning, and we denote the task set as $\mathcal{T}$. 
 {During training}, several tasks are sampled from $\mathcal{T}$ in order to train an adaptive model which can perform well on new tasks after  {test}. Different samples are fetched in the selected task, and are split into two sets, the support set and the query set. 
To be specific, when a task $\mathcal{T}_{i}$ is sampled from the task set $\mathcal{T}$, it is trained through the corresponding  loss function $\mathcal{L}_{\mathcal{T}_{i}}$ to update parameters.
\textit{$K_{s}$} samples (the support set $\mathcal{D}_{\mathcal{T}_{i}}^{s}$) are fetched from $\mathcal{T}_{i}$ for training using $\mathcal{L}_{\mathcal{T}_{i}}$, and then tested on \textit{$K_{q}$} new samples (the query set $\mathcal{D}_{\mathcal{T}_{i}}^{q}$) sampled from $\mathcal{T}_{i}$ each time.
The model is optimized according to the errors of all the samples in query set calculated by $\mathcal{L}_{\mathcal{T}_{i}}$, that is to say, the test error of the query set is the meta-training error of  {training}.
During  {test}, new tasks will be sampled, trained on its corresponding support sets, and then tested on query sets to demonstrate the meta-performance.

\begin{figure}[t]
	
	\centering
	\subfigure[The inner and outer loop updates of MAML.]{
		\includegraphics[width=0.45\textwidth]{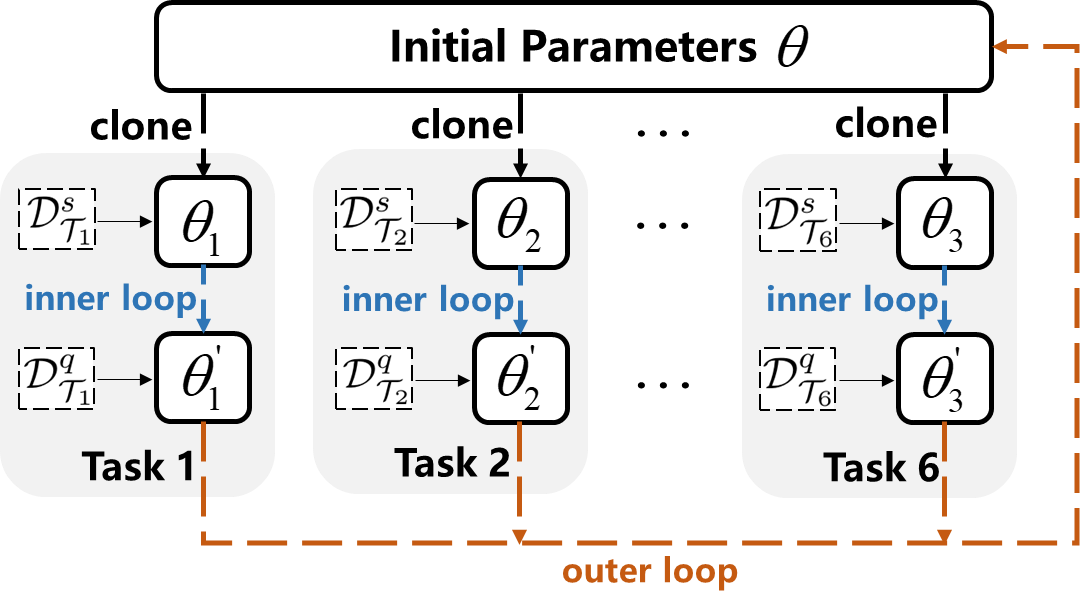}
		\label{update_maml}
	}
	\quad
	\subfigure[The parameter update in our adversarial task.]{
		\includegraphics[width=0.45\textwidth]{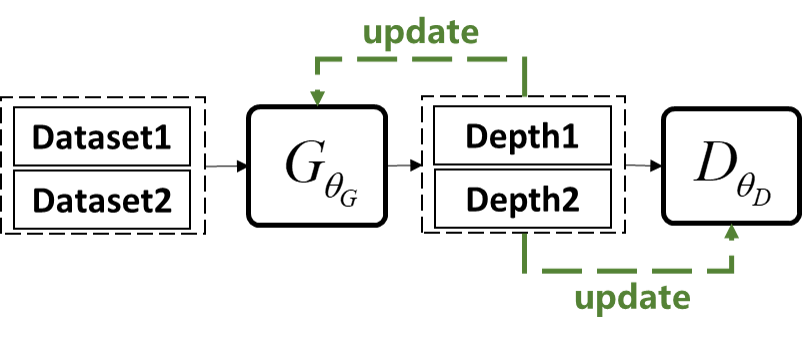}
		\label{update_depth}
	}
	\caption{Parameter update of the proposed framework. The parameters of the generator and the discriminator update alternately.}
	\label{update}
\end{figure}

MAML~\cite{finn2017model} is designed to train the initial parameters of a model, so that the model can adapt to a new task through one or several gradient updates  given a handful of data from the new task. As shown in Figure \ref{update_maml}, considering a model $\textit{f}_{\theta}$ with parameters $\theta$, the model optimizes through training on different tasks (six tasks in this work). For each task $\mathcal{T}_{i}$, the model updates its parameters $\theta$ to $\theta'_{i}$. The model is updated through one or several gradient descents. The update using one gradient descent is as follow: 
\begin{equation}
\label{meta_gradient}
\begin{aligned}
\theta'_{i} = \theta - \alpha \nabla_{\theta}\mathcal{L}_{\mathcal{T}_{i}}(\textit{f}_{\theta})\;,
\end{aligned}
\end{equation}
where $\alpha$ is the learning rate of the fast adaptation. Then, the model is trained by minimizing all the errors of $\mathcal{D}_{\mathcal{T}_{i}}^{q}$ tested on the updated model $\textit{f}_{\theta'_{i}}$. The meta-optimization is performed through 
stochastic gradient descent (SGD)  according to the losses of query sets and the initial parameters $\theta$ are updated as follows:
\begin{equation}
\label{meta_update}
\begin{aligned}
\theta \leftarrow \theta - \beta \nabla_{\theta}\sum_{\mathcal{T}_{i}\sim p(\mathcal{T})}\mathcal{L}_{\mathcal{T}_{i}}(\textit{f}_{\theta'_{i}})\;,
\end{aligned}
\end{equation}
where $\beta$ is the learning rate of the meta-learning.


\begin{figure*}[h]
	\centering
	\includegraphics[width=0.98\textwidth]{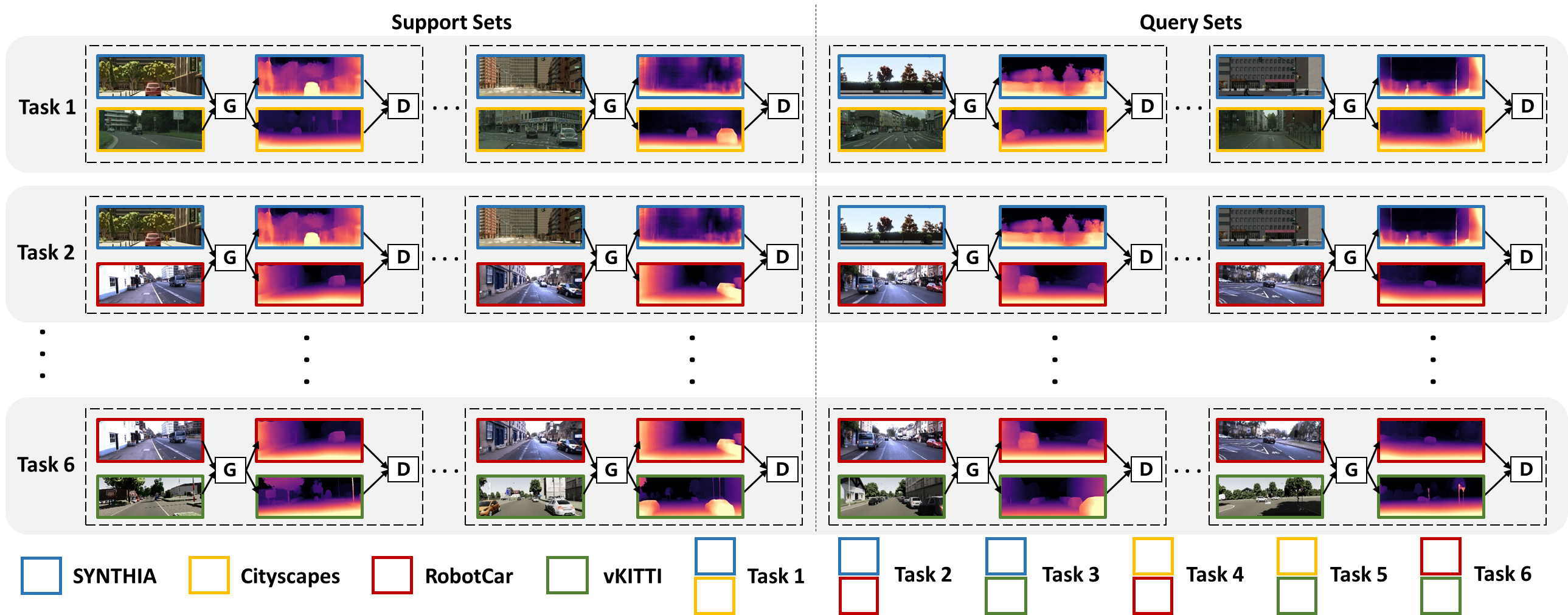}
	\caption{The framework of our adversarial  {meta-learning} monocular depth estimation method. Four datasets are used to construct six datasets pairs for meta-learning.}
	\label{network}
\end{figure*}


In this work, the single task for meta-learning is our designed adversarial depth estimation task introduced in sub-section \ref{sec:adversarial}, as shown in Figure \ref{UDA}.
In the  {training} phase, a task-wise support set $\mathcal{D}_{\mathcal{T}_{i}}^{s}$ and a task-wise query set $\mathcal{D}_{\mathcal{T}_{i}}^{q}$ are sampled from the sampled task $\mathcal{T}_{i}$, as shown in Figure \ref{update_maml}.
The depth estimation model is trained through the support set $\mathcal{D}_{\mathcal{T}_{i}}^{s}$, and for each training sample in  $\mathcal{D}_{\mathcal{T}_{i}}^{s}$, the parameters of the model are fast updated in the inner loop. The initial parameters of the  model are updated with the outer loop.
We denote the generator and the discriminator in our model as $G_{\theta_{G}}$ and  $D_{\theta_{D}}$, parameterized by $\theta_{G}$ and $\theta_{D}$, respectively. 
As illustrated in Figure  \ref{update_maml}, the \textbf{inner loop} of our meta-learning based method updates the model parameters in each separate task, and the \textbf{outer loop} updates initial model parameters of   the inner loop.
The loss function for inner loop training is the combination of depth estimate loss $\mathcal{L}_{\mathcal{T}_{i}}^{G}$ and the adversarial  loss $\mathcal{L}_{\mathcal{T}_{i}}^{D}$. Then, the inner loop update loss $\mathcal{L}_{\mathcal{T}_{i}}^{in}(G_{\theta_{G}}, D_{\theta_{D}})$ for task  $\mathcal{T}_{i}$ is:
\begin{equation}
\label{meta_inner}
\left\{
\begin{aligned}
&\mathcal{L}_{\mathcal{T}_{i}}^{in}(G_{\theta_{G}}) = \mathcal{L}_{\mathcal{T}_{i}}^{G}\big(G_{\theta_{G}}(\mathcal{D}_{\mathcal{T}_{i}}^{s})\big) \\
&\mathcal{L}_{\mathcal{T}_{i}}^{in}(D_{\theta_{D}}) =  \mathcal{L}_{\mathcal{T}_{i}}^{D}\Big(D_{\theta_{D}}\big(G_{\theta_{G}}(\mathcal{D}_{\mathcal{T}_{i}}^{s})\big)\Big) \;.
\end{aligned}
\right.
\end{equation}

\begin{algorithm}[t]
	\caption{Adversarial Domain-Adaptive Depth Estimation} 
	\label{alg:algorithm1}	
	\KwIn{$\mathcal{T}$: task set} 
	\KwIn{$\alpha$, $\beta$: step size hyper-parameters} 	
	Initialize parameters $\theta$ 	
	\While{not done} 
	{ 
		Sample batch of tasks $\mathcal{T}_{i}\sim \mathcal{T}$		
		\For{each $\mathcal{T}_{i}$} 
		{ 
			Sample $K_s$ and $K_q$ datapoints from $\mathcal{T}_{i}$ as support set $\mathcal{D}_{\mathcal{T}_{i}}^{s}$ and query set $\mathcal{D}_{\mathcal{T}_{i}}^{q}$
			
			Evaluate $\nabla_{\theta}\mathcal{L}_{\mathcal{T}_{i}}^{in}(G_{\theta_{G}})$ using $\mathcal{D}_{\mathcal{T}_{i}}^{s}$ and $\mathcal{L}_{\mathcal{T}_{i}}^{G}$ in Equation \ref{meta_inner}
			
			Evaluate $\nabla_{\theta}\mathcal{L}_{\mathcal{T}_{i}}^{in}(D_{\theta_{D}})$ using $\mathcal{D}_{\mathcal{T}_{i}}^{s}$ and $\mathcal{L}_{\mathcal{T}_{i}}^{D}$ in Equation \ref{meta_inner}
			
			Compute adapted parameters with gradient descent: $\theta_{{G}_{i}}' = \theta_{G} - \alpha \nabla_{\theta_{G}}\mathcal{L}_{\mathcal{T}_{i}}^{in}(G_{\theta_{G}})$, $\theta_{{D}_{i}}' = \theta_{D} - \alpha \nabla_{\theta_{D}}\mathcal{L}_{\mathcal{T}_{i}}^{in}(D_{\theta_{D}})$ 
			
		}
		
		Update $\theta_{G} \leftarrow \theta_{G} - \beta \nabla_{\theta_{G}}\sum_{\mathcal{T}_{i}\sim p(\mathcal{T})}\mathcal{L}_{\mathcal{T}_{i}}^{out}(G_{\theta_{G}'})$, $\theta_{D} \leftarrow \theta_{D} - \beta \nabla_{\theta_{D}}\sum_{\mathcal{T}_{i}\sim p(\mathcal{T})}\mathcal{L}_{\mathcal{T}_{i}}^{out}(D_{\theta_{D}'})$  using  $\mathcal{D}_{\mathcal{T}_{i}}^{q}$, and $\mathcal{L}_{\mathcal{T}_{i}}^{u}$, $\mathcal{L}_{\mathcal{T}_{i}}^{D}$ in Equation \ref{meta_outer}
	} 	
\end{algorithm}

The parameters of $G_{\theta_{G}}$ and $D_{\theta_{D}}$ are updated alternately. 
For each task $\mathcal{T}_{i}$, $\mathcal{L}_{\mathcal{T}_{i}}^{in}(G_{\theta_{G}})$ and  $\mathcal{L}_{\mathcal{T}_{i}}^{in}(D_{\theta_{D}})$ are calculated to deduce their gradient, respectively. Then, the parameters $\theta_{{G}}$ and $\theta_{{D}}$ are updated to $\theta_{{G}_{i}}'$ and $\theta_{{D}_{i}}'$ with gradient descent. Note that  the parameters of the generator and discriminator are updated in an alternative manner, as shown in Figure \ref{update_depth}. The outer loop update is conducted when all the inner loops are terminated, and the parameters are optimized by minimizing the loss generated by $G_{\theta_{{G}_{i}}'}$ and $D_{\theta_{{D}_{i}}'}$ with query set  $\mathcal{D}_{\mathcal{T}_{i}}^{q}$. The optimization objective of the outer loop meta-update (meta-objective) is as follows: 
\begin{equation} 
\label{meta_outer} 
\left\{
\begin{aligned}
&\mathcal{L}_{\mathcal{T}_{i}}^{out}(G_{\theta_{{G}_{i}}'}) =  \mathcal{L}_{\mathcal{T}_{i}}^{u}\big(G_{\theta_{{G}_{i}}'}(\mathcal{D}_{\mathcal{T}_{i}}^{q})\big) \\
&\mathcal{L}_{\mathcal{T}_{i}}^{out}(D_{\theta_{{D}_{i}}'}) =  \mathcal{L}_{\mathcal{T}}^{D}\Big(D_{\theta_{{D}}}\big(G_{\theta_{{G}_{i}}'}(\mathcal{D}_{\mathcal{T}_{i}}^{q})\big)\Big) \;.
\end{aligned}
\right.
\end{equation}

Then, the corresponding gradients can be calculated by Equation \ref{meta_outer} to update the model parameters. The parameters updating process of our method is illustrated in Figure \ref{update} and is summarized in Algorithm \ref{alg:algorithm1}. The  framework of our adversarial meta-training monocular depth estimation is shown in Figure \ref{network}.
During  {test}, only the inner loop updates the parameters for a specific dataset and the outer loop is switched off. That is, we use  initial parameters  trained during  {training} and continue to train a model when given a new dataset during  {test phase}.

\subsection{Cross-task depth consistency}

\begin{figure}[t]
	\centering
	\includegraphics[scale=0.5]{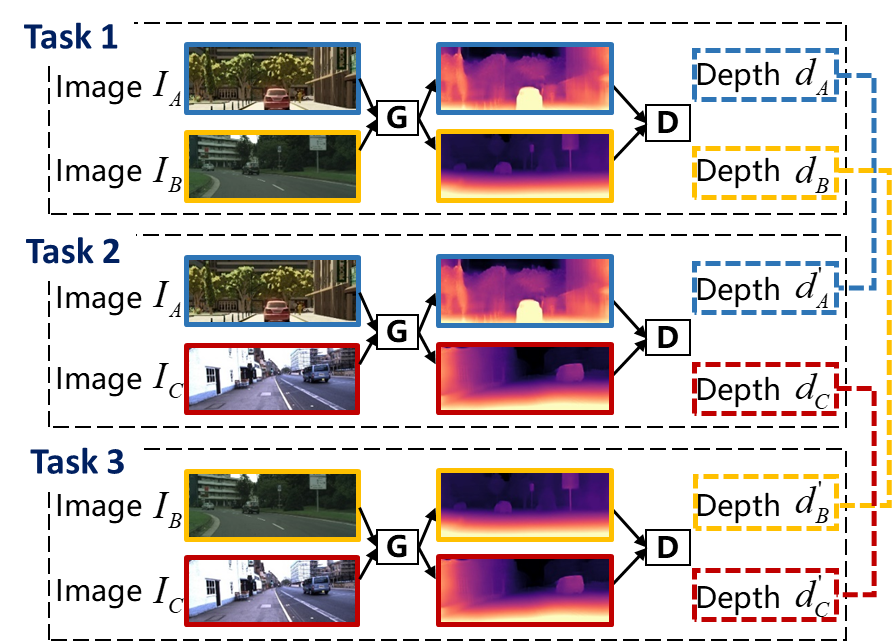}
	\caption{The illustration of our cross-task depth consistency. The depth maps estimated from the same scene (the depth maps in the dotted boxes with the same color) in different tasks are forced to be identical (connected with dotted lines).}
	\label{consistent}
\end{figure}

In the proposed framework, an image is sent to different tasks for depth estimation during  {training} and an example consisting of three different tasks is shown in  Figure \ref{consistent}. As an image (Image $I_{A}$ for example) is sent to different tasks (Task 1 and Task 2) for estimation, the predicted depth maps ($d_{A}$ and $d_{A}'$) can be different due to the differences in network parameters. Nevertheless, the depth map of a specific scene ought to be the same and the cross-task depth consistency constraint is proposed to compel the depth estimated in different tasks to be identical. 
The differences in network parameters of different tasks are caused by the individual parameter updates in inner loops with different training samples.  
The  cross-task depth consistency is added in the outer loop update and denoted as $\mathcal{L}_{c}$, which is calculated by:
\begin{equation}
\label{cross_domain}
\begin{aligned}
\mathcal{L}_{c} = |d_{I_{i}}-d'_{I_{i}}|\;,
\end{aligned}
\end{equation}
where $d_{I_{i}}$ and  $d_{I_{i}}'$ are the estimated depth maps  in different tasks.
Since our $\mathcal{L}_{c}$ compels the depth estimated in different tasks to be identical, it can smoothen the  convergence curve and improve the performance of our method. Related experiments can be found in Section \ref{sec:experiments}.
Then, the optimization objectives of the generator and discriminator for outer loop meta-update can be updated as:
\begin{equation} 
\label{meta_outer1} 
\left\{
\begin{aligned}
&\mathcal{L}_{\mathcal{T}_{i}}^{out}(G_{\theta_{{G}_{i}}'}) =  \mathcal{L}_{\mathcal{T}_{i}}^{u}\big(G_{\theta_{{G}_{i}}'}(\mathcal{D}_{\mathcal{T}_{i}}^{q})\big) + \mathcal{L}_{c} \\
&\mathcal{L}_{\mathcal{T}_{i}}^{out}(D_{\theta_{{D}_{i}}'}) =  \mathcal{L}_{\mathcal{T}}^{D}\Big(D_{\theta_{{D}}}\big(G_{\theta_{{G}_{i}}'}(\mathcal{D}_{\mathcal{T}_{i}}^{q})\big)\Big) \;.
\end{aligned}
\right.
\end{equation}
Equation \ref{meta_outer} in Algorithm \ref{alg:algorithm1} should also be updated as Equation \ref{meta_outer1}.

\section{Experiments}
\label{sec:experiments}

%

\subsection{Datasets}
Four datasets, SYNTHIA~\cite{ros2016synthia}, Cityscapes~\cite{cordts2016cityscapes}, Oxford RobotCar~\cite{maddern20171} and vKITTI~\cite{gaidon2016virtual}, are used for  {training}. To prove the effectiveness of our method, three different datasets: KITTI~\cite{KITTI}, nuScenes~\cite{caesar2020nuscenes}  and DrivingStereo~\cite{yang2019drivingstereo}, are used for  {test} to prove that our method can generalize well to different new scenes.



\noindent{\textbf{CityScapes}~\cite{cordts2016cityscapes}:
	CityScapes is a well-known benchmark comprised of images collected in 50 cities 
	in Germany, including 
	a collection of 22,973 stereo pairs with the resolution of 1024$\times$2048.}

\noindent{\textbf{Oxford RobotCar}~\cite{maddern20171}: Oxford RobotCar is composed of image sequences collected in different weather conditions  with the resolution of 960$\times$1280. In our experiments, we use the sequences from 2014-12-09-13-21-02 for training.
}

\noindent{\textbf{vKITTI}~\cite{gaidon2016virtual}: Virtual KITTI (vKITTI) contains 21,260 image-depth pairs with the  resolution of 375$\times$1242, which is generated by computer graphic engines. vKITTI tries to imitate the style of KITTI and generates images under different environmental conditions.}

\noindent{\textbf{SYNTHIA}~\cite{ros2016synthia}:
	SYNTHIA is a synthetic dataset  containing images generated in the style of different seasons, scenarios, weathers, and  illuminations.}

\noindent{\textbf{KITTI}~\cite{KITTI}: KITTI dataset is a common benchmark in computer vision and contains  42,382 rectified stereo pairs in the resolution of 375$\times$1242. Though KITTI contains the ground truth of depth which is collected by Radar, we only use the images.}

\noindent{\textbf{nuScenes}~\cite{caesar2020nuscenes}: NuSenses dataset comprises 1000 image sequences collected in street scenes with the  resolution of 900$\times$1600. Different sensors are equipped, including 6 cameras deployed.
}

{\noindent{\textbf{DrivingStereo}~\cite{yang2019drivingstereo}: DrivingStereo is a large-scale stereo dataset which contains  more than 180,000 images collected from diverse driving scenarios. The resolution of the stereo pairs is 800$\times$1762  and all the images have their corresponding ground truth depth collected by LiDAR.
}

\subsection{Implementation Details}

Our framework is implemented in Pytorch~\cite{paszke2017automatic} and optimized by Adam~\cite{Adam}, which includes a  generator and a discriminator.
The generator adopts the depth estimation architecture in ~\cite{2019digging} as the basic model. The discriminator is the same as the discriminator in CycleGAN~\cite{zhu2017unpaired}. We use MAML~\cite{finn2017model} as the pipeline of  meta-learning.
During  {training}, we select $K_s$ = 4, $K_q$ = 4, $\alpha$ = $10^{-4}$, $\beta$ = $10^{-4}$ in Algorithm \ref{alg:algorithm1} for the training of the fast adapters in inner loop and the meta-adapter in outer loop. The batch size is set to be six, which means that all the  tasks in  task set are chosen for training in our domain-adaptive depth estimation algorithm.
During  {test}, we reload the network parameters obtained in  {training} as the initial parameters.

Since the training images in different datasets have different resolutions and some of them contains some superfluous detail (e.g., car hood), we crop the images and resize them to the  resolution of  640$\times$192 in  {training}.   
 {During test}, we resize the images in KITTI into the resolution of  640$\times$192, and resize the images in  nuScenes and DrivingStereo into the resolution of  640$\times$320.
As the proposed method aims to obtain appropriate initial parameters that can adapt rapidly to unseen domains, the quantity of network parameters remain the same.


\subsection{Ablation Experiments}

\begin{table*}
	\footnotesize
	\caption{The evaluation of each component used in the proposed algorithm  on 
		Eigen split~\cite{Eigen2014Depth} and the proposed strategies are proved to be effective.  }	
	\begin{center}
		\begin{tabular}{|c|cccc|ccc|}
			\hline
			\multirow{2}{*}{Method} & Abs Rel & Sq Rel & RMSE &RMSE log & $\delta<1.25$ &
			$\delta<1.25^{2}$ & $\delta<1.25^{3}$\\
			\cline{2-8}
			{}&\multicolumn{4}{c}{Lower is better} &\multicolumn{3}{|c|}{Higher is better}\\  
			\hline
			\multicolumn{8}{|c|}{\textbf{ {Training} (5 epochs)}}   \\     
			\hline
			Basic &   0.183  &   2.908  &   6.366  &   0.252  &   0.781  &   0.928  &   0.968  \\			
			MAML & 0.172  &   1.794  &   6.415  &   0.258  &   0.766  &   0.920  &   0.965   \\
			MAML + AT &   0.163  &   1.542  &   5.824  &   0.238  &   0.788  &   0.933  &   0.972  \\
			Ours (MAML + AT +   $\mathcal{L}_{c}$) &   \textbf{0.149 } &  \textbf{1.280} &   \textbf{5.624 } &   \textbf{0.224}  &   \textbf{0.807}  &   \textbf{0.942}  &   \textbf{0.976}     \\
			\hline
			\multicolumn{8}{|c|}{\textbf{ {Test} (0.5 epoch)}}   \\ 
			\hline
			Basic&   0.121  &   0.930  &   4.917  &   0.196  &   0.864  &   0.958  &   0.981 \\	
			MAML & 0.121  &   0.861  &   4.820  &   0.196  &   0.860  &   0.958  &   0.982\\	
			MAML + AT & 0.118  &   0.829  &   4.791  &   0.195  &   0.866  &   0.959  &   0.982  \\
			Ours (MAML + AT +   $\mathcal{L}_{c}$) &   \textbf{0.115}  &   \textbf{0.784}  &   \textbf{4.612}  &   \textbf{0.191}  &   \textbf{0.872}  &   \textbf{0.961}  &   \textbf{0.983}  \\
			\hline 				
			
		\end{tabular}
		
	\end{center}
	\label{tab:ablation}
\end{table*}

\begin{figure}[t]	
	\centering
	\subfigure[ {Train.}]
	{\includegraphics[width=0.5\textwidth]{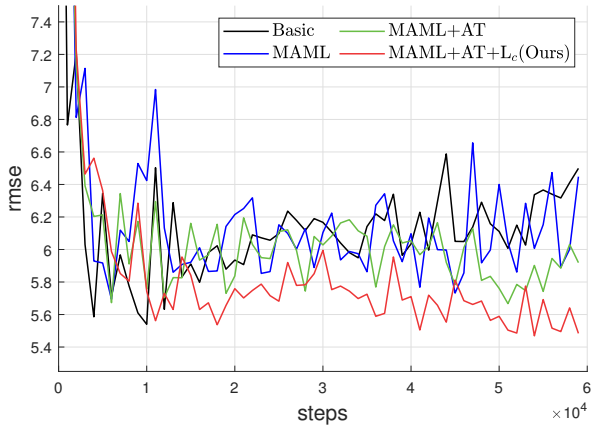}
		\label{fig:meta-training}
	}
	\quad	
	\subfigure[ {Test.}]
	{\includegraphics[width=0.5\textwidth]{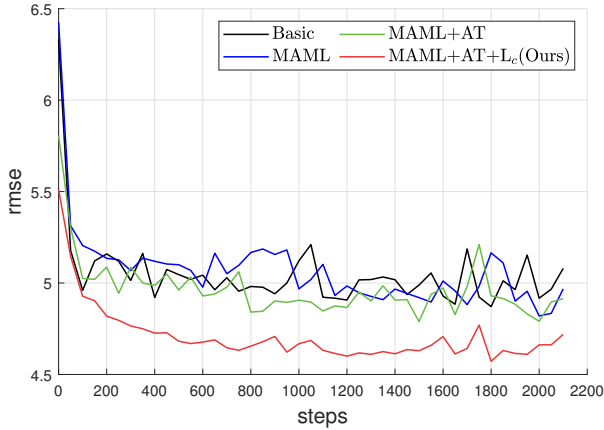}
		\label{fig:meta-testing}
	}
	\caption{The learning curves of different models in  {training and test}  phase on Eigen test-split~\cite{Eigen2014Depth} of KITTI dataset.}
	\label{curves}
	\vspace{-10pt}
\end{figure}


Ablation experiments are conducted on the KITTI dataset to analyze if the strategies used in our method are effective. The quantitative results are shown in Table \ref{tab:ablation} and the learning curves of different models are plotted in Figure \ref{curves}. 
The \textbf{Basic} in Table \ref{tab:ablation} refers to the depth estimation model in~\cite{2019digging}. We employ the MAML pipeline, our designed adversarial depth estimation task, the proposed cross-task depth consistency constraint on the basic model separately to analyze if mechanisms
in our method can improve the transferability of depth estimation.
In the basic method, we train the depth estimation model in~\cite{2019digging} with all the four datasets.  
\textbf{MAML} indicates that the adversarial training task designed is not used and the model is trained in the pipeline of MAML with the basic model. 
\textbf{AT} refers to that we use the \textbf{A}dversarial \textbf{T}ask 
designed in Figure \ref{UDA} for  {training}.
Since our cross-task depth consistency constraint $\mathcal{L}_{c}$ is designed for the outer loop update in meta-learning, it can only be employed along with MAML. 
Our method use both AT and $\mathcal{L}_{c}$ in  {training}.

During  {training}, four datasets are used to train initial parameters for  {the test phase}. We test the trained model directly on Eigen test-split~\cite{Eigen2014Depth} without  {test} to evaluate if the  {training} progress achieves convergence. The performances of different variants of our model  are listed in Table \ref{tab:ablation} ( {Training} (5 epochs)) and the corresponding training curves are plotted in Figure \ref{fig:meta-training}. 
From Figure \ref{fig:meta-training}, we can find that nearly all the modified models perform better than the basic one on the average. However, the performance of the model trained with MAML directly is instable and does not perform significantly better than the basic model, which may result from the meta-overfitting. Fortunately, the adversarial task  designed helps to smoothen the training curves, which proves that our modified meta-learning algorithm helps to overcome the meta-overfitting. Further, our  proposed $\mathcal{L}_{c}$ accelerates convergence and improves the performance as well.   

\begin{table*}
	
		\caption{Comparisons of the proposed method ~\cite{2019digging} on KITTI, nuScenes and DrivingStereo datasets. Our algorithm trained for few iterations achieves comparable results with ~\cite{2019digging} trained for much more iterations on different datasets. }
		\begin{center}
			\resizebox{\textwidth}{!}{
				\begin{tabular}{|c|c|c|c|cccc|ccc|}
					\hline
					\multirow{2}{*}{Dataset} &\multirow{2}{*}{Method} &  \multirow{2}{*}{Resolution} &  \multirow{2}{*}{Training time}  
					& Abs Rel & Sq Rel & RMSE &RMSE log & $\delta<1.25$ & $\delta<1.25^{2}$ & $\delta<1.25^{3}$\\
					\cline{5-11}
					{}&{}&{}&{}&\multicolumn{4}{c}{Lower is better} &\multicolumn{3}{|c|}{Higher is better}\\
					\hline
					\multirow{2}{*}{KITTI} &Monodepth2~\cite{2019digging} & 640$\times$192& 60K interations & \textbf{0.115} &0.903 &4.863& 0.193& \textbf{0.877} &0.959 &0.981  \\
					{}&Ours &640$\times$192 & 1.5K interations& \textbf{0.115}  &   \textbf{0.784}  &   \textbf{4.612}  &   \textbf{0.191}  &   0.872  &   \textbf{0.961}  &   \textbf{0.983} \\				
					\hline
					\multirow{2}{*}{nuScenes} &Monodepth2~\cite{2019digging} & 640$\times$320 &40K interations&  0.122 & 1.380 & \textbf{6.368} & 0.222 & 0.852 & \textbf{0.940} & 0.968 \\
					{}&Ours &640$\times$320 & 3K interations&  \textbf{0.118} & \textbf{1.306} & 6.620 & \textbf{0.221} & \textbf{0.853} & 0.936 & \textbf{0.969} \\				
					\hline
					\multirow{2}{*}{DrivingStereo} &Monodepth2~\cite{2019digging} &640$\times$320 & 40K interations&  0.097 & 0.825 & 5.148 & 0.141 & \textbf{0.916} & \textbf{0.980} & 0.991 \\
					{}&Ours &640$\times$320 & 3K interations  & \textbf{0.093} & \textbf{0.801} & \textbf{5.058} & \textbf{0.139} & 0.912 & \textbf{0.980} & \textbf{0.993} \\	 
					\hline   
				\end{tabular}
			}
		\end{center}	
		\label{tab:other_datasets}
\end{table*}


\begin{table*}
	\caption{Comparisons of our method and some state-of-the-art works. Our algorithm trained for few updates achieves comparable results with other works trained for dozens of epoches. The maximum value of depth is 80m. }
	\begin{center}
		\resizebox{\textwidth}{!}{
			\begin{tabular}{|c|c|c|c|cccc|ccc|}
				\hline
				\multirow{2}{*}{Method} & \multirow{2}{*}{Supervision}& \multirow{2}{*}{Training time}  &  \multirow{2}{*}{Resolution } 
				& Abs Rel & Sq Rel & RMSE &RMSE log & $\delta<1.25$ & $\delta<1.25^{2}$ & $\delta<1.25^{3}$\\
				\cline{5-11}
				{}&{}&{}&{}&\multicolumn{4}{c}{Lower is better} &\multicolumn{3}{|c|}{Higher is better}\\
				\hline
				Eigen \textit{et al.}~\cite{Eigen2014Depth} & Depth & /& 612$\times$184 & 0.190 & 1.515 & 7.156 & 0.270 & 0.692 & 0.899 & 0.967 \\
				Godard \textit{et al.}~\cite{Godard2017Unsupervised} & Pose  &50 epochs& 512$\times$256& 0.148 & 1.344 & 5.927 & 0.247 & 0.803 & 0.922 & 0.964 \\
				SfM-Learner~\cite{Zhou2017Unsupervised} & No  & 150K iterations &128$\times$416& 0.183 & 1.595 & 6.709 & 0.270 & 0.734 & 0.902 & 0.959 \\
				CC~\cite{CC}	& No & 100K iterations & 832$\times$256 &  0.140 &	1.070 &	5.326 &	0.217 &	0.826 & 0.941 &	0.975\\
				Alex \textit{et al.}~\cite{2019bilateral}	& No  & 50 epochs &512$\times$256&  0.133 &	1.126 &	5.515 &	0.231 &	0.826 & 0.934 &	0.969\\
				Bian \textit{et al.}~\cite{bian2019unsupervised}	& No & 200K iterations &832$\times$256 &0.137 &	1.089 &	5.439 &	0.217 &0.830  & 0.942 &	0.975\\
				GASDA~\cite{zhao2019geometry} & No &40 epochs & 640$\times$192 & 0.149 &1.003 &	4.995 &	0.227 &0.824  & 0.941 &	0.973\\	
				SynDeMo~\cite{bozorgtabar2019syndemo} & No &280K iterations & 608$\times$160& 0.112 &0.740 &	4.619 &	0.187 &0.863  & 0.958 &	0.983\\	
				Monodepth2~\cite{2019digging} & No &20 epochs & 1024$\times$320& 0.115 &0.882 &	4.701 &	0.190 &0.879  & 0.961 &	0.982\\	
				Zhao \textit{et al.}~\cite{zhao2020towards}  & No  &50 epochs& 832$\times$256& 0.113 &{0.704} &	4.581 &	{0.184} &0.871 & 0.961 &	{0.984}\\								
				Johnston \textit{et al.}~\cite{johnston2020self} (ResNet18)& No & 20 epochs& 640$\times$192& 0.111   &0.941& 4.817  & 0.189 & 0.885 &  0.961 &  0.981 \\

				\hline
				Ours (ResNet18) & No & \textbf{0.5 epoch} (1.5K iterations) & 640$\times$192&  0.115  &   0.784  &   4.612  &   0.191  &   0.872  &   0.961  &   0.983\\	 
				\hline   
		\end{tabular}}
	\end{center}
	
	\label{tab:kitti_results}
\end{table*}

\begin{figure*}[t]
	\begin{center}   
		\includegraphics[width=0.95\textwidth]{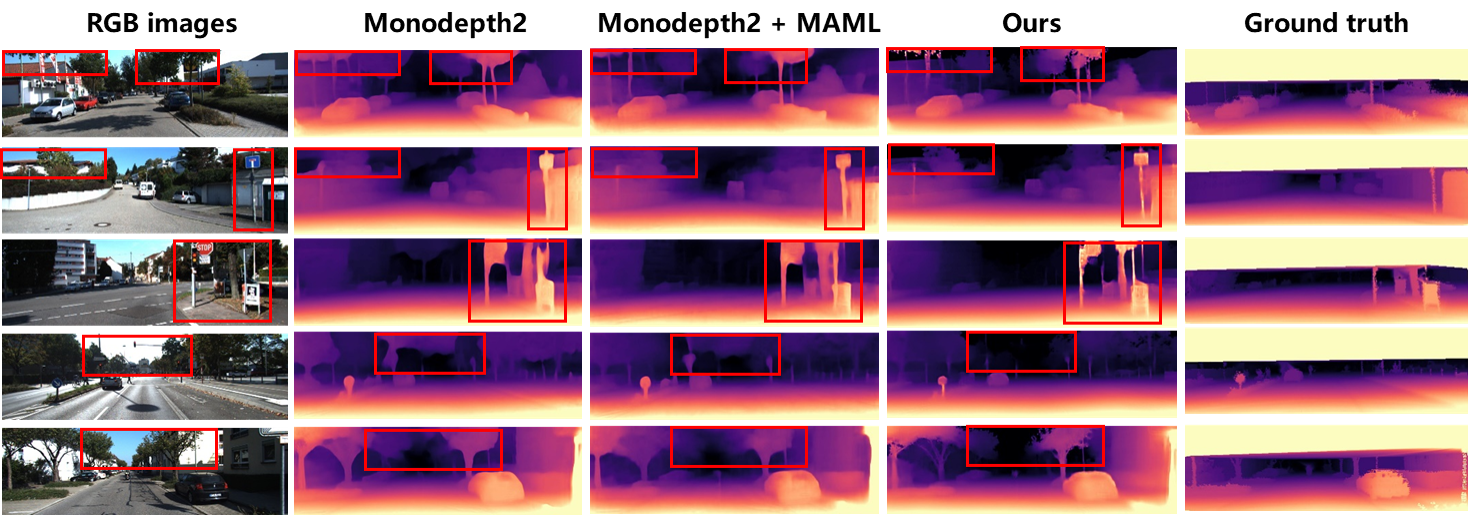}
	\end{center}
	\caption{Comparisons of the depth estimation results of different methods, including the original ~\cite{2019digging}, ~\cite{2019digging} trained in  the pipeline of basic MAML, and the proposed method.  }
	\label{fig:depth}
\end{figure*}

\begin{figure*}[t]
	\begin{center}   
		\includegraphics[width=1\textwidth]{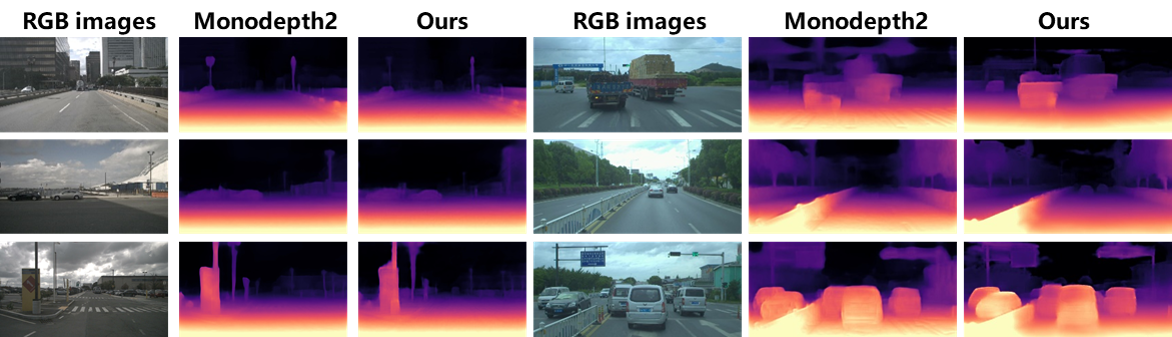}
	\end{center}
	\caption{Comparisons of the predicted depth maps using the proposed method and ~\cite{2019digging} on nuScenes (left) and Drivingstereo (right) datasets.}
	\label{fig:other_datasets}
\end{figure*}

Then, the test process  is conducted on KITTI. We continue to train the model obtained in  {training} on Eigen train-split~\cite{Eigen2014Depth} and then test the model on Eigen test-split. The results are shown in Table \ref{tab:ablation} ( {Test} (0.5 epoch)) and Figure \ref{fig:meta-testing}. 
As shown in Figure \ref{fig:meta-testing}, the performances of different models in  {test} are similar to those in  {training}. The MAML training pipeline fails to improve the transferability of the basic model notably and our designed adversarial task helps to reduce the negative effect of  {training}.
Since our proposed cross-task depth consistency constraint $\mathcal{L}_{c}$ compels the predicted depth maps of the same scene in different tasks to be identical, the structure information can be learnt. Thus,  $\mathcal{L}_{c}$ helps to  smoothen the training curve and improve the performance significantly. 
According to Figure \ref{fig:meta-testing}, both our designed adversarial task and the proposed $\mathcal{L}_{c}$ improve the performance and convergence rate.
All the results demonstrate that our method can generalize well to KITTI, which is new to the training process.

\subsection{Transferability of Our Depth Estimation Model}

To demonstrate the generalizability of the proposed method, we conduct experiments on different auto-driving datasets: KITTI~\cite{KITTI}, nuScenes~\cite{caesar2020nuscenes}  and DrivingStereo~\cite{yang2019drivingstereo}. As the backbone of our depth estimation network is \cite{2019digging}, we compare our work with \cite{2019digging} and the quantitative comparison results are shown in Table \ref{tab:other_datasets}. The results prove that our method obtains satisfactory results on several new datasets with limited training iterations  when compared to \cite{2019digging} trained with much more iterations.  
Since KITTI is a widely used dataset in depth estimation, 
we compare our methods with some other state-of-the-art methods~\cite{Eigen2014Depth,Godard2017Unsupervised,Zhou2017Unsupervised,CC,2019bilateral,bian2019unsupervised,2019digging,zhao2020towards,johnston2020self,zhao2019geometry,bozorgtabar2019syndemo}, as shown in Table \ref{tab:kitti_results}.  
As for nuScenes and DrivingStereo, few related works conduct depth estimation experiments on them, thus we only compare our work with ~\cite{2019digging}. 


For KITTI, we adopt the Eigen split~\cite{Eigen2014Depth} used in the compared works in  {test phase} to evaluate the transferability of our method. We compare different methods with ours and the quantitative 
results are shown in Table \ref{tab:kitti_results}. 
It should be noticed that all the competing methods have access to KITTI dataset during the whole training. 
On the contrary, our model is trained on other datasets in  {training} first and then trained on KITTI for limited steps, and obvious domain gaps exist between these datasets and KITTI. 
As shown in Table \ref{tab:kitti_results}, the compared methods are  trained on KITTI for dozens of epochs, while our method is trained on KITTI for about 0.5 epoch during  {test}. Actually, as shown in Figure \ref{fig:meta-testing}, our method achieves a satisfactory performance after 500 iterations (1/6 epoch).
The experimental results show that our method performs comparable to state-of-the-art works through a few updates with a relative low resolution, which proves the outstanding transferability of our methods.
Figure \ref{fig:depth} visualizes the depth maps estimated by different models, and we compare the model trained by Monodepth2~\cite{2019digging} in the basic MAML with our method to prove the effectiveness of our proposed strategies further.
Our method specializes in capturing and reconstructing the detail of objects and provides sharper predictions, for example, the outline of trees. The segmentation of the background is also more distinct and we highlight some regions in Figure \ref{fig:depth}. 
As our algorithm is not sensitive to model structure, other methods, such as~\cite{zhao2020towards,johnston2020self} can be used as the baseline of our algorithm, which can improve the performance of depth estimation further.

For nuScenes, we use about 200,000 images for training and validation, and randomly choose 660 images for test. 
For DrivingStereo, 350,000 images are utilized for training and validation, and 547 images are selected stochastically to test. 
We compare our method with the state-of-the-art depth estimation method Monodepth2~\cite{2019digging} to prove the generalizability of our method on these two datasets. Similar to the experiments on KITTI dataset, \cite{2019digging} have access to the whole training set during training process. On the contrary,  our method is trained for limited iterations, which means that our method only sees some part images in the training set for just one time. The quantitative and qualitative results are shown in Table \ref{tab:other_datasets} and Figure \ref{fig:other_datasets}. 
The experimental results show that our method performs well on KITTI, nuScenes and DrivingStereo, which demonstrates that our method adapts well to new datasets rapidly.

\section{Conclusions and Future Works}

To deal with the issue of poor generalization, we propose an adversarial domain-adaptive algorithm for depth estimation, which can transfer well to new, unseen datasets in the presented work.
We design an adversarial task for monocular depth  estimation and the task is trained in the manner of meta-learning. The proposed adversarial task alleviates the issue of overfitting in meta-learning when the training tasks are limited.
In addition, a cross-task depth consistency constraint is  imposed for meta-update.  
Experiments show that our method generalizes well to new, unseen datasets, which demonstrates that our method learns to adapt for depth estimation gracefully. 
In the presented work, our method is trained and tested on autonomous driving datasets. In the future, we will dedicate to strengthen the transferability of our method further, for example, to enable our method to adapt well to not only outdoor scenes, but also indoor scenes.

%
%

\ifCLASSOPTIONcaptionsoff
  \newpage
\fi



%

{\small
\bibliographystyle{IEEEtran}
\bibliography{IEEEexample}
}




%

%






\end{document}